# Model-Free Renewable Scenario Generation Using Generative Adversarial Networks
Yize Chen, Yishen Wang, Daniel Kirschen and Baosen Zhang

*Abstract*—Scenario generation is an important step in the operation and planning of power systems with high renewable penetrations. In this work, we proposed a data-driven approach for scenario generation using generative adversarial networks, which is based on two interconnected deep neural networks. Compared with existing methods based on probabilistic models that are often hard to scale or sample from, our method is data-driven, and captures renewable energy production patterns in both temporal and spatial dimensions for a large number of correlated resources. For validation, we use wind and solar time-series data from NREL integration data sets. We demonstrate that the proposed method is able to generate realistic wind and photovoltaic power profiles with full diversity of behaviors. We also illustrate how to generate scenarios based on different conditions of interest by using labeled data during training. For example, scenarios can be conditioned on weather events (e.g. high wind day, intense ramp events or large forecasts errors) or time of the year (e,g. solar generation for a day in July). Because of the feedforward nature of the neural networks, scenarios can be generated extremely efficiently without sophisticated sampling techniques.

## I. INTRODUCTION

High levels of renewables penetration pose challenges in the operation, scheduling, and planning of power systems. Since renewables are intermittent and stochastic, accurately modeling the uncertainties in them is key to overcoming these challenges [1], [2]. One widely used approach to capture the uncertainties in renewable resources is by using a set of time-series scenarios [3]. By using a set of possible power generation scenarios, renewables producers and system operators are able to make decisions that take uncertainties into account, such as stochastic economic dispatch/unit commitment, optimal operation of wind and storage systems, and trading strategies (e.g., see [4], [5], [6], [7] and the references within).

Despite the tremendous advances recently achieved, scenario generation remains a challenging problem [8], [9]. The dynamic and time-varying nature of weather, the nonlinear and bounded power conversion processes, and the complex spatial and temporal interactions make model-based approaches difficult to apply and hard to scale, especially when multiple renewable power plants are considered. These models are typically constructed based on statistical assumptions that may not hold or difficult to test in practice, and sampling from high-dimensional distributions (e.g. non-Gaussian) is also nontrivial [3]. In addition, some of these methods depend on certain probabilistic forecasts as inputs, which may limit the diversity of the generated scenarios and under-explore the overall variability of renewable resources.

To overcome these difficulties, in this work, we propose a data-driven (or model-free) approach by adopting *generative* methods. Specifically, we propose to utilize the power of the recently discovered machine learning concept of *Generative Adversarial Networks* (GANs) [10] to fulfill the task of scenario generation. Generative models have become a research frontier in computer vision and machine learning area, with the promise of utilizing large volumes of unlabeled training data. There are two key benefits of applying such class of methods. The first is that they can directly generate new scenarios based on historical data, without explicitly specifying a model or fitting probability distributions. The second is that they use unsupervised learning, avoiding cumbersome manual labellings that are sometimes impossible for large datasets. In the image processing community, GANs are able to generate realistic images that are of far better quality compared to other methods [10], [11], [12].

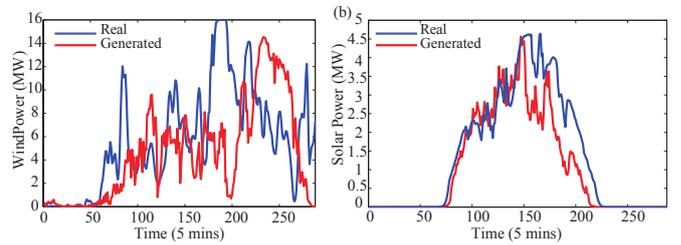

Fig. 1. Group of historical scenarios versus generated scenarios using our method for wind (left) and solar (right) power generation. Blue curves are true historical data and red curves are generated scenarios. Both scenarios exhibit rapid variation and strong diurnal patterns that are hallmarks of wind and solar power.

### A. Literature Review

Past works have focused mainly on model-based methods that first find probabilistic distributions then sample from these distributions to generate renewable power scenarios [8]. Many families of probabilistic models have been proposed in previous research. For example, in [13], [14], [15], [4], [16], [17], copula methods are first applied to model the distribution and correlation of forecast errors, then scenarios are generated either via Monte Carlo Sampling (MCS) or Latin Hypercube sampling (LHS). In [18], an empirical cumulative distribution function is used to replace copula estimation to model the uncertainty of wind power. While in [19], a generalized dynamic factor model (GFDM) is adopted to preserve the correlation structure between load and wind power scenarios. However, it is difficult to capture the temporal dynamics

Y. Chen, D. Kirschen and B. Zhang authors are with the Department of Electrical Engineering at the University of Washington, emails: {yizechen,kirschen,zhangbao}@uw.edu. Y. Wang is with GEIRI North America, email: yishen.wang@geirina.net

of renewable generation by first and second order statistics alone. In addition, the difficulty of obtaining good long-term forecasts also prevent copula methods from generating realistic scenarios.

Another popular class of scenario generation methods make use of time series. In [20], a first-order autoregressive time-series model with increasing noise is applied to approximate the behavior of forecast errors, while in [3] autoregressive moving average (ARMA) model is used to generate spatiotemporal scenarios with given power generation profiles at each renewables generation site. In [21], the author translate the autoregressive model into a state space form, so that the dependencies can be structurally analyzed more directly. Although simple to implement, autoregressive model and state-space specifications are prone to overfitting and misidentification of patterns. Capturing enough of the diversity in the renewable generation processes can also be difficult using these models due to potentially the need to include a large number of states.

Recently, several machine learning algorithms are also proposed for scenario generation. In [22], a Radial Basis Function Neural Networks (RBFNN) is coupled with particle swarm optimization (PSO) algorithm to generate scenarios with input from numerical weather predictions (NWP). In [23], [24], neural network models are trained to output either time-series power generation or occurrence probability. Compared to copula or time series methods, these machine learning based algorithms may potentially better capture the nonlinear dynamics of renewable generation processes, but all of these depend on careful selection of input features and is nontrivial to tune and use in practice.

In summary, most of the above methods first fit a model using historical observations, and then the fitted probabilistic models are sampled to generate new scenarios. Some of these methods may also require pre-processing of data. Despite the significant advances, scenario generation remains a challenging problem. The dynamic and time-varying nature of weather, the nonlinear and bounded power conversion processes, and the complex spatial and temporal interactions make model-based approaches difficult to apply and hard to scale. A single set of model parameters normally cannot capture these complex dynamics, especially when multiple renewable power plants are considered. These models are typically constructed based on statistical assumptions that may not hold or difficult to test in practice (e.g., forecast errors are Gaussian). Sampling from high-dimensional distributions (e.g. non-Gaussian) is also nontrivial [3]. In addition, methods like Gaussian copula and ARMA depend on certain probabilistic forecasts as inputs. The spatiotemporal relations and accuracy of the forecast directly affect the diversity of the generated scenarios.

*B. Proposed Method and Main Contributions*

In this paper, we show that GANs can also effectively generate renewable scenarios, with suitable modifications that takes into account the fact that renewable resources are driven by physical processes and have different characteristics compared to images.

Fig. 1 shows examples of our generated daily scenarios with a comparison to historical scenarios. These generated scenarios correctly capture the rapid variations and strong diurnal cycles in wind and solar. Note we explicitly chose examples where the historical data and the generated scenarios do not match each other perfectly. Our goal is to generate *new and distinct* scenarios that capture the intrinsic features of the historical data, but not to simply memorize the training data. More examples are shown later in the paper (Fig. 4), and we conduct a host of tests to show that the generated scenarios have the same visual and statistical properties as historical data.

The intuition behind GANs is to leverage the power of deep neural networks (DNNs) to both express complex nonlinear relationships (the generator) as well as classify complex signals (the discriminator). The key insight of GAN is to set up a minimax two player game between the generator DNN and the discriminator DNN (thus the use of "adversarial" in the name). During each training epoch, the generator updates its weights to generate "fake" samples trying to "fool" the discriminator network, while the discriminator tries to tell the difference between true historical samples and generated samples. In theory, at reaching the Nash equilibrium, the optimal solution of GANs will provide us a generator that can exactly recover the distribution of the real data so that the discriminator would be unable to tell whether a sample came from the generator or from the historical training data. At this point, generated scenarios are indistinguishable from real historical data, and are thus as realistic as possible. Fig. 2 shows the general architecture of a GANs' training procedure under our specific setting.

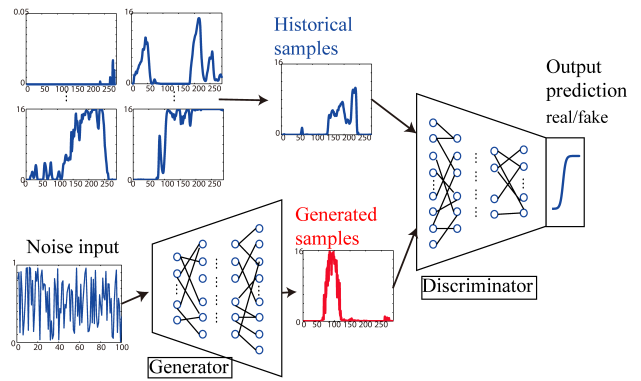

Fig. 2. The architecture for GANs used for wind scenario generation. The input to the generator is noise that comes from an easily sampled distribution (e.g., Gaussian), and the generator transforms the noise through the DNN with the goal of making its output to have the same characteristics as the real historical data. The discriminator has two sources of inputs, the "real" historical data and the generator data, and tries to distinguish between them. After training is completed, the generator can produce samples with the same distribution as the real data, without efforts to explicitly model this distribution.

The main contributions of this paper are:
1) *Data-driven scenario generation*: We propose a model-free, data-driven and scalable approach for renewables scenario generation. By employing generative adversarial networks, we can generate scenarios which capture the spatial and temporal correlations of renewable power plants. To our knowledge, this is the first work applying

2) *Conditional scenario generations*: We enable generation of scenarios of specific characteristics (e.g., high wind days, seasonal solar outputs) by using a simple label in the training process. This procedure could be easily adjusted to capture different conditions of interest.
3) *Efficient algorithms*: We show GANs can be trained with little or no manual adjustments, and it can be swiftly scaled up to generate large and diverse set of renewable profiles.

All of the code and data described in this paper are publicly available at https://github.com/chennnnnyize/Renewables_Scenario_Gen_GAN. The rest of the paper is organized as follows: Section II rigorously formulates the mathematical problems; Section III proposes and describes the GANs model; results are illustrated and evaluated in Section IV; and Section V concludes the paper.

## II. PROBLEM FORMULATION

In this section, we give the mathematical formulations for three scenario generation tasks of interests: 1) scenario generation for single renewable resource; 2) scenario generation for multiple correlated renewable resources; and 3) scenario generation conditioned on different events.

### A. Single Time-Series Scenario Generation

Consider a set of historical data for a group of renewable resources at $N$ sites. For site $j$, let $\mathbf{x}_j$ be the vector of historical data indexed by time, $t = 1, \ldots, T$, and $j$ ranges from 1 to $N$. Our objective is to train a generative model based on GANs by utilizing historical power generation data $\{\mathbf{x}_j\}, j = 1, \ldots, N$ as the training set. Generated scenarios should be capable of describing the same stochastic processes as training samples and exhibiting a variety of different modes representing all possible variations and patterns seen during training.

### B. Scenario Generation for Multiple Sites

In a large system, multiple renewable resources needs to be considered at the same time. Here we are interested in simultaneously generating multiple scenarios for a given group of geographical close sites. We have historical power generation observations $\{\mathbf{x}_j\}, j = 1, \ldots, N$ for $N$ sites of interests with the same time horizon. The generated scenarios should capture both the temporal and spatial correlations between the resources, as well as the marginal distribution of each individual resource.

In some situations a point forecast is given and scenarios should be thought as the forecasting error. Our approach can be easily applied by simply replacing the training samples with the historical forecast errors. Based on different forecasting technologies, there may or may not be correlations among the errors. Our approach would automatically generate statistically correct scenarios without any explicit assumptions.

### C. Event-Based Scenario Generation

In addition to the standard scenario generation process described above, we may want to generate scenarios with distinct properties. For instance, an operator may be interested in scenarios that capture the solar output of a hot summer day. We incorporate these given properties into the training process by labeling each training samples with an assigned label to represent the event. Specifically, we use a label vector $\mathbf{y}$ to classify and record certain properties in an observation $\mathbf{x}_j$.

Thus in this part we are interested in scenario generation conditioned on the label $y$, while samples having same label should follow the similar properties. Our objective here is to train a generative model based on GANs using historical conditional power generation data $\{\mathbf{x}_j | y_j\}, j = 1, \ldots, N$ as a training set.

## III. GENERATIVE ADVERSARIAL NETWORKS

In this section, we introduce the GANs [10] and how they are adapted to our applications of interests for renewables scenario generation. We first review the method and formulate the objectives as well as the loss functions, then describe how to incorporate additional information into the model training process.

### A. GANs with Wasserstein Distance

The architecture of GANs we use is shown in Fig. 2. Assume observations $x_j^t$ for times $t \in T$ of renewable power are available for each power plant $j$, $j = 1, \ldots, N$. Let the true distribution of the observation be denoted by $\mathbb{P}_X$, which is of course unknown and hard to model. Suppose we have access to a group of noise vector input $z$ under a known distribution $Z \sim \mathbb{P}_Z$ that is easily sampled from (e.g., jointly Gaussian). Our goal is to transform a sample $z$ drawn from $\mathbb{P}_Z$ such that it follows $\mathbb{P}_X$ (without ever learning $\mathbb{P}_X$ explicitly). This is accomplished by simultaneously training two deep neural networks: the generator network and the discriminator network. Let $G$ denote the generator function parametrized by $\theta^{(G)}$, which we write as $G(\cdot; \theta^{(G)})$; Let $D$ denote the generator function parametrized by $\theta^{(D)}$, which we write as $D(\cdot; \theta^{(D)})$. Here, $\theta^{(G)}$ and $\theta^{(D)}$ are the weights of two neural networks, respectively. For convenience, we sometimes suppress the symbol $\theta$.

*Generator*: During the training process, the generator is trained to take a batch of inputs and by taking a series of up-sampling operations by neurons of different functions to output realistic scenarios. Suppose that $Z$ is a random variable with distribution $\mathbb{P}_Z$. Then $G(Z; \theta^{(G)})$ is a new random variable, whose distribution we denote as $\mathbb{P}_G$.

*Discriminator*: The discriminator is trained simultaneously with the generator. It takes input samples either coming from real historical data or coming from generator, and by taking a series of operations of down-sampling using another deep neural network, it outputs a continuous value $p_{real}$ that measures to what extent the input samples belong to $\mathbb{P}_X$. The discriminator can be expressed as

$$p_{real} = D(x; \theta^{(D)}) \quad (1)$$

where $x$ may come from $\mathbb{P}_{data}$ or $\mathbb{P}_Z$. The discriminator is trained to learn to distinguish between $\mathbb{P}_X$ from $\mathbb{P}_G$, and thus to maximize the difference between $\mathbb{E}[D(X)]$ (real data) and $[D(G(Z))]$ (generated data).

With the objectives for discriminator and generator defined, we need to formulate loss function $L_G$ for generator and $L_D$ for discriminator to train them (i.e., update neural networks' weights based on the losses). In order to set up the game between $G$ and $D$ so that they can be trained simultaneously, we also need to construct a game's value function $V(G,D)$. During training, a batch of samples drawn with distribution $\mathbb{P}_Z$ are fed into the generator. At the same time, a batch of real historical samples are fed into the discriminator. A small $L_G$ shall reflect the generated samples are as realistic as possible from the discriminator's perspective, e.g., the generated scenarios are looking like historical scenarios for the discriminator. Similarly, a small $L_D$ indicates discriminator is good at telling the difference between generated scenarios and historical scenarios, which reflect there is a large difference between $\mathbb{P}_G$ and $\mathbb{P}_X$. Following this guideline and the loss defined in [25], we can write $L_D$ and $L_G$ as followed:

$$L_G = -\mathbb{E}_Z[D(G(Z))] \qquad (2a)$$
$$L_D = -\mathbb{E}_X[D(X)] + \mathbb{E}_Z[D(G(Z))]. \qquad (2b)$$

Since a large discriminator output means the sample is more realistic, the generator will try to minimize the expectation of $-D(G(\cdot))$ by varying $G$ (for a given $D$), resulting in the loss function in (2a). On the other hand, for a given $G$, the discriminator wants to minimize the expectation of $D(G(\cdot))$, and the same time maximizing the score of real historical data. This gives the loss function in (2b). Note the functions $D$ and $G$ are parametrized by the weights of the neural networks.

We then combine (2a) and (2b) to form a two-player minimax game with the value function $V(G,D)$:

$$\min_{\theta^{(G)}} \max_{\theta^{(D)}} V(G,D) = \mathbb{E}_X[D(X)] - \mathbb{E}_Z[D(G(Z))] \qquad (3)$$

where $V(G,D)$ is the negative of $L_D$.

At early stage of training, $G$ just generates scenario samples $G(z)$ totally different from samples in $\mathbb{P}_X$, and discriminator can reject these samples with high confidence. In that case, $L_D$ is small, and $L_G$, $V(G,D)$ are both large. The generator gradually learns to generate samples that could let $D$ output high confidence to be true, while at the same time the discriminator is also trained to distinguish these newly fed generated samples from $G$. As training moves on and goes near to the optimal solution, $G$ is able to generate samples that look as realistic as real data with a small $L_G$ value, while $D$ is unable to distinguish $G(z)$ from $\mathbb{P}_X$ with large $L_D$. Eventually, we are able to learn an unsupervised representation of the probability distribution of renewables scenarios from the output of $G$.

More formally, the minimax objective (3) of the game can be interpreted as the dual of the so-called Wasserstein distance (Earth-Mover distance) [26]. Given two random variables $X$ and $Y$ with marginal distribution $f_X$ and $f_Y$, respectively, let $\Gamma$ denote the set of all possible joint distributions that has marginals of $f_X$ and $f_Y$. Wasserstein distance between them is defined as

$$W(X,Y) = \inf_{f_{XY} \in \Gamma} \int |x-y| f_{XY}(x,y) dx dy. \qquad (4)$$

This distance, although technical, measures the effort (or "cost") needed to transport the probability distribution of $X$ to the probability distribution of $Y$: the inf in (4) finds the joint distribution that gets $x$ and $y$ to have smallest distance while maintaining the marginals [27]. The connection to GANs comes from the fact that we are precisely trying to get two random variables, $\mathbb{P}_X(D(X))$ and $\mathbb{P}_Z(D(G(X)))$, to be close to each other. It turns out that

$$W(D(X), D(G(Z))) = \sup_{\theta^{(D)}} \{\mathbb{E}_X[D(X)] - \mathbb{E}_Z[D(G(Z))]\}, \qquad (5)$$

where the expectations can be computed as empirical means.

As is shown in Fig. 3b, once the Wasserstein distance we estimate using 3 stops decreasing or reaches pre-set limits, the "cost" of transforming a generated sample to original sample has been minimized. So the distance between the distribution of $G(Z)$ and $\mathbb{P}_X$ is minimized. Thus we find the optimal generator $G^*$. In the GANs community, there is a growing body of literature about the choice of loss functions. Here we chose to use the Wasserstein distance [25] instead of the original Jensen-Shannon divergence proposed in [10]. This is because Wasserstein distance directly calculates the distance between two distributions $\mathbb{P}_G$ and $\mathbb{P}_X$. Since we want to generate scenarios that reflect the variability of renewables generation processes, training using Wasserstein distance allows us to capture all of the modes in training samples which are all coming from $\mathbb{P}_X$. Training using Jensen-Shannon divergence tends to lead the generator to generate a single pattern of power profile that has the highest probability. In this paper, we want to generate scenarios that reflect diverse modes of renewables, which is accomplished by using the Wasserstein distance to directly measure the distance between distributions of real historical data and generated samples.

*B. Conditional GANs*

In an unconditioned generative model, we do not control the specific types of the samples being generated by $G$. Sometimes, we are interested in scenarios "conditioned on" certain class of events, e.g., calm days with intermittent wind, or windy days with farms at full load capacity.

Conditional generation is done by incorporating more information to the training procedure of GANs, such that the generated samples conforming to same properties as certain class of training samples. Inspired by supervised learning where we have labels for each training samples, here we propose to combine event labels with training samples, and thus the objective for $G$ is to generate samples under given class [11]. More formally, the problem can be written as:

$$\min_{\theta^{(G)}} \max_{\theta^{(D)}} = \mathbb{E}_X[D(X|y)] - \mathbb{E}_Z[D(G(Z|y))], \qquad (6)$$

where $y$ encodes different type of classes of conditions.

Class labels are assigned based on user-defined classification metrics, such as the mean of daily power generation values, the month of training samples coming from, etc. Note that such class labels are just representation of samples' events reflected by the power generation data distribution, GANs should be able to learn the conditional distribution and generate samples based on any given meaningful classification metric. We will show the results in Section. IV with given labels based on mean values as well as seasonal information. Since such conditional GANs is only modifying the unconditional GANs model proposed in Section. III-A with a label vector input, both models can be trained using Algorithm 1.

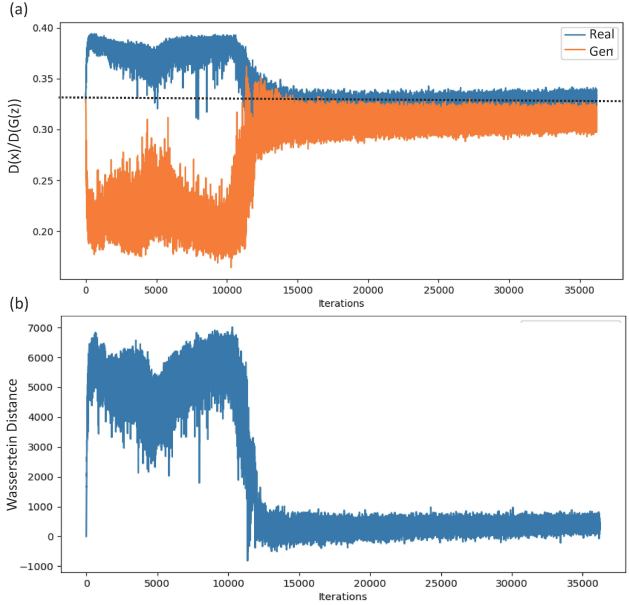

Fig. 3. Training evolution for GANs on a wind dataset. (a) The outputs from the discriminator $D(x)/D(G(z))$ during training illustrates the evolution of generated samples $G(z)$. At the start, the generated samples (orange) and the real samples (blue) are easily distinguished at the discriminator. As training progresses, they are increasing difficult to distinguish. (b) The empirical Wasserstein distance between the distribution of the real sample and the generated samples, where close to zero means that the two distributions are close to each other.

In our setup, $D(x; \theta^{(D)})$ and $G(z; \theta^{(G)})$ are both differentiable functions which contain different neural layers composed of multilayer perceptron, convolution, normalization, max-pooling and Rectified Linear Units (ReLU). Thus we can use standard training methods (e.g., gradient descent) on these two networks to optimize their performances. Training is implemented in a batch updating style, while a learning rate self-adjustable gradient descent algorithm *RMSProp* is applied for weights updates in both discriminator and generator neural networks [28]. Clipping is also applied to constrain $D(x; \theta^{(D)})$ to satisfy certain technical conditions as well as preventing gradients explosion [25]. Detailed model structures and training procedure are described in Section. IV.

## IV. RESULTS

In this section we illustrate our algorithm on several different setups for wind and solar scenario generation. We first show that the generated scenarios are visually indistinguish-

**Algorithm 1** Conditional GANs for Scenario Generation

**Require:** Learning rate $\alpha$, clipping parameter $c$, batch size $m$, Number of iterations for discriminator per generator iteration $n_{discri}$

**Require:** Initial weights $\theta^{(D)}$ for discriminator and $\theta^{(G)}$ for generator

  **while** $\theta^{(D)}$ has not converged **do**
    **for** $t = 0, ..., n_{discri}$ **do**
      *# Update parameter for Discriminator*
      Sample batch from historical data:
      $\{(x^{(i)}, y^{(i)})\}_{i=1}^{m} \, \mathbb{P}_X$
      Sample batch from Gaussian distribution:
      $\{z^{(i)}, y^{(i)}\}_{i=1}^{m} from \, \mathbb{P}_Z$
      Update discriminator nets using gradient descent:
      $g_{\theta^{(D)}} \leftarrow \nabla_{\theta^{(D)}} [-\frac{1}{m} \sum_{i=1}^{m} D(x^{(i)}|y^{(i)}) + \frac{1}{m} \sum_{i=1}^{m} D(G(z^{(i)}|y^{(i)}))]$
      $\theta^{(D)} \leftarrow \theta^{(D)} - \alpha \cdot RMSProp(\theta^{(D)}, g_{\theta^{(D)}})$
      $\theta^{(D)} \leftarrow clip(w, -c, c)$
    **end for**
    *# Update parameter for Generator*
    Update generator nets using gradient descent:
    $g_{\theta^{(G)}} \leftarrow \nabla_{\theta^{(G)}} \frac{1}{m} \sum_{i=1}^{m} D(G(z^{(i)}|y^{(i)}))$
    $\theta^{(G)} \leftarrow \theta^{(G)} - \alpha \cdot RMSProp(\theta^{(G)}, g_{\theta^{(G)}})$
  **end while**

able from real historical samples, then we show that they also exhibit the same statistical properties [29], [30]. These results suggest that using GANs would provide an efficient, scalable, and flexible approach for generating high-quality renewable scenarios.

### A. Data Description

We build training and validation dataset using power generation data from NREL Wind[1] and Solar[2] Integration Datasets [31]. The original data has resolution of 5 minutes. We choose 24 wind farms and 32 solar power plants located in the State of Washington to use as the training and validating datasets. We shuffle the daily samples and use 80% of them as the training data, and the remaining 20% as the testing datasets Along with the wind read measurements, we also collect the corresponding 24-hour ahead forecast data, which is later used for conditional generation based on forecasts error. The 10% and 90% quantile forecasts are also available for Gaussian copula method setup. All of these power generation sites are of geographical proximity which exhibit correlated (although not completely similar) stochastic behaviors. Our method can easily handle joint generation of scenarios across multiple locations by using historical data from these locations as inputs with no changes to the algorithm. Thus the spatiotemporal relationships are learned automatically.

### B. Model Architecture and Details of Training

The architecture of our deep convolutional neural network is inspired by the architecture of DCGAN and Wasserstein

---
[1] https://www.nrel.gov/grid/wind-integration-data.html
[2] https://www.nrel.gov/grid/sind-toolkit.html

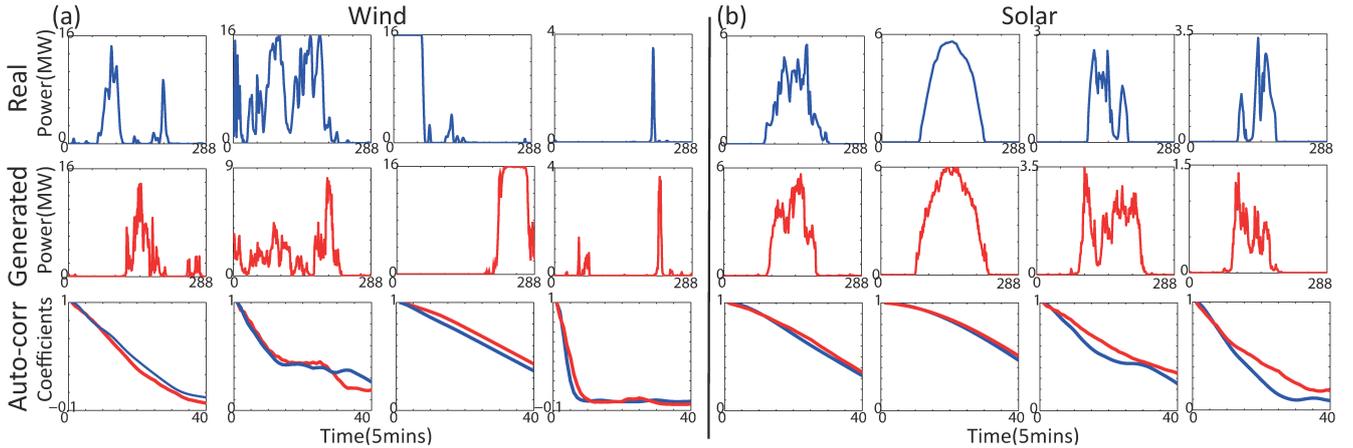

Fig. 4. Selected samples from our validation sets (top) versus generated samples from our trained GANs (middle) for both wind and solar groups. The pair of samples are selected using Euclidean distance based search. Without using these validation samples in training, our GANs is able to generate samples with similar behaviors and exhibit a diverse range of patterns. The autocorrelation plots (bottom) also verify generated samples' ability to capture the correct time-correlations.

GAN[12], [25]. The generator $G$ includes 2 de-convolutional layers with stride size of $2 \times 2$ to firstly up-sample the input noise $z$, while the discriminator $D$ includes 2 convolutional layers with stride size of $2 \times 2$ to down-sample a scenario $x$. The generator starts with fully connected multilayer perceptron for upsampling. The discriminator has a reversed architecture with a single sigmoid output. We observe two convolution layers are adequate to represent the daily dynamics for the training set, and is efficient for training. Details of the generator and discriminator model parameters are listed in the Table I. Note that both $G$ and $D$ are realized as DNN, which can be programmed and trained via standard open source platforms such as Tensorflow [32]. All the program for GANs model is implemented in Python platform with two Nvidia Titan GPUs to accelerate the deep neural networks' training procedure.

TABLE I
The GANs model structure. MLP denotes the multilayer perceptron followed by number of neurons; Conv/Conv_transpose denotes the convolutional/deconvolutional layers followed by number of filters; Sigmoid is used to constrain the discriminator's output in $[0,1]$.

|  | Generator $G$ | Discriminator $D$ |
| --- | --- | --- |
| **Input** | 100 | 24*24 |
| **Layer 1** | MLP, 2048 | Conv, 64 |
| **Layer 2** | MLP, 1024 | Conv, 128 |
| **Layer 3** | MLP, 128 | MLP, 1024 |
| **Layer 4** | Conv_transpose, 128 | MLP, 128 |
| **Layer 5** | Conv_transpose, 64 |  |

All models in this paper are trained using RmsProp optimizer with a mini-batch size of 32. All weights for neurons in neural networks were initialized from a centered Normal distribution with standard deviation of 0.02. Batch normalization is adopted before every layer except the input layer to stabilize learning by normalizing the input of every layer to have zero mean and unit variance. With exception of the output layer, ReLU activation is used in the generator and Leaky-ReLU activation is used in the discriminator. In all experiments, $n_{discri}$ was set to 4, so that the model were training alternatively between 4 steps of optimizing $D$ and 1 step of $G$. We observed model convergence in the loss for discriminator in all the group of experiments. Once the discriminator has converged to similar outputs value for $D(G(z))$ and $D(x)$, the generator was able to generate realistic power generation samples.

We also set up Gaussian copula method for scenario generation in order to compare the result with proposed method [13], [16]. In order to capture the interdependence structure, we recursively estimated the Gaussian copula $\Sigma \in \mathbb{R}^{d \times d}$ based on $d-$dimension historical power generation observations $\{\mathbf{x_j}\}$, $j = 1, ..., N$ for $N$ sites of interests. Then with a Normal random number generator with zero mean and covariance matrix $\Sigma$, we are able to draw a group of scenarios (after passing through the Copula function).

### C. Scenario Generation

We firstly trained the model to validate that GANs can generate scenario with diurnal patterns. The training evolution for this training dataset is shown in Fig. 3.

For the first $10,000$ iterations, the output of discriminator has a large difference between generated and real historical samples, which indicates that the discriminator can easily distinguish between the sources of the samples. After $15,000$ iterations of training, the Wasserstein distance $\mathbb{E}_{x \sim \mathbb{P}_{data}}[D(x)] - \mathbb{E}_{z \sim \mathbb{P}_Z}[D(G(z))]$ shown in Fig. 3(b) already converged to near 0. We keep the training until $35,000$ iterations to demonstrate the training procedure is stable and once converged, the empirical distribution of a batch of generated scenarios are very close to the empirical distribution of a batch of training scenarios. Thus we are getting to the optimal solution.

We then fed the trained generator with $2,500$ noise vectors $Z$ drawn from the pre-defined Gaussian distribution $z \sim \mathbb{P}_Z$. Some generated samples are shown in Fig. 4 with comparison to some samples from the validation set. We see that the generated scenarios closely resemble scenarios from the validation set, which were not used in the training of the GANs. Next, we show that generated scenarios have two important properties:

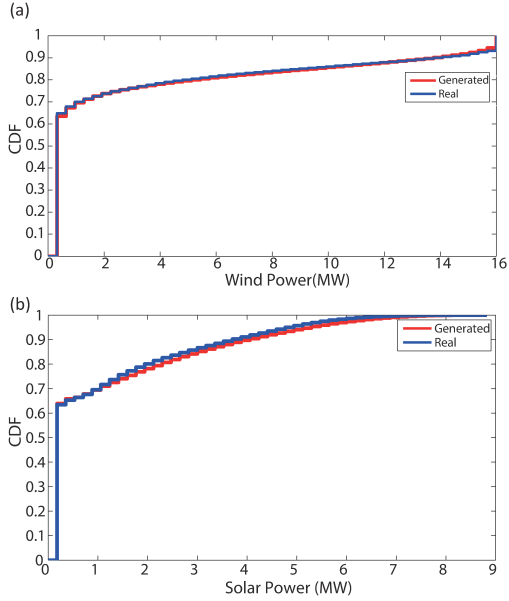

Fig. 5. We evaluate the quality of generated scenarios by calculating the marginal CDFs of generated and historical scenarios of wind (Figure (a)) and solar (Figure (b)), respectively. The two CDFs are nearly identical for both solar and wind.

1) *Mode Diversity*: The diversity of modes variation are well captured in the generated scenarios. For example, the scenarios exhibit hallmark characteristics of renewable generation profiles: e.g., large peaks, diurnal variations, fast ramps in power, etc. For instance, in the third column in Fig. 4, the validating and generated sample both include sharp changes in its power. Using a traditional model-based approach to capture all of these characteristics would be challenging, and may require significant manual effort.

2) *Statistical Resemblance*: We also verify that generated scenarios has the same statistical properties as the historical data. For the original and generated samples shown in Fig. 4, we first calculate and compare sample autocorrelation coefficients $R(\tau)$ with respect to time-lag $\tau$:

$$R(\tau) = \frac{E[(S_t - \mu)(S_{t+\tau} - \mu)]}{\sigma^2} \qquad (7)$$

where $S$ represents the stochastic time-series of either generated samples or historical samples with mean $\mu$ and variance $\sigma$. Autocorrelation represents the temporal correlation at a renewable resource, and capture the correct temporal behavior is of critical importance to power system operations. The bottom rows of Fig. 4 verify that the real-generated pair have very similar autocorrelation coefficients.

In addition to comparing the stochastic behaviors in single time-series, in Fig.5 we show the cumulative distribution function (CDF) of historical validating samples and GANs-generated samples. We find that the two CDFs nearly lie on top of each other. This indicates the capability of GANs to generate samples with the correct marginal distributions.

PSD evaluates the spectral energy distribution that would be found per unit time. To verify the periodic component and temporal correlation of each individual scenario, we calculate the PSD ranging from 6 days to 2 hours for the validation set and generated scenarios. In Fig. 6 we plot the results for wind scenarios and solar scenarios respectively. We observe that for both cases, generated scenarios closely follow the overall shape of historical observations coming from the validation set.

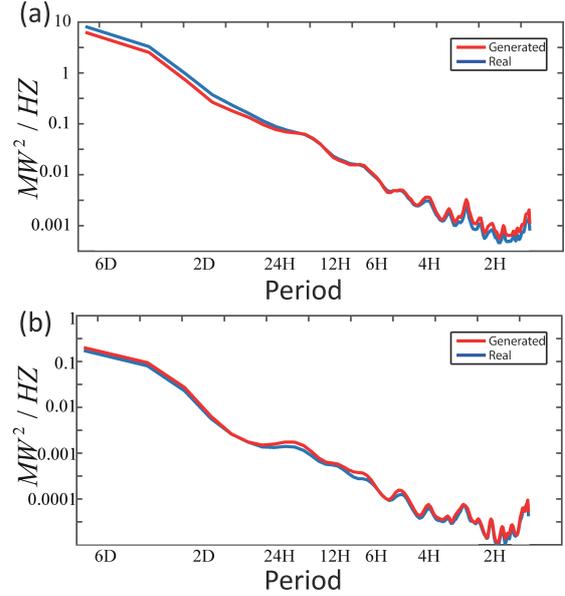

Fig. 6. The power spectral density(PSD) plots for both wind (Fig. 6a) and solar (Fig. 6b) scenario generation.

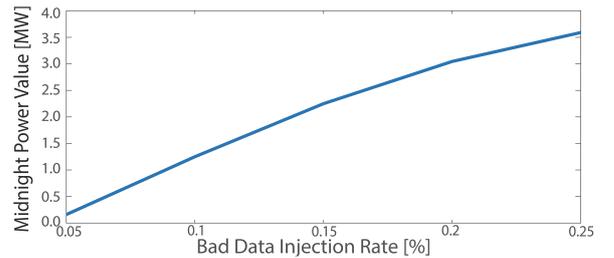

Fig. 7. The solar power generation from 12am-1am versus the rate of wind samples injected into solar samples training set.

To test the robustness of our data-driven model against bad training data, we simulate the injection of wind power generation samples into solar power generation samples. Both the bad data and clean training data are normalized into $[0, 1]$ scale. Since solar generation samples have daily patterns with no power output during night, here we plot the midnight power generation value with respect to the rate of injected wind samples in Fig. 7. There is a trend that with more wind samples injected, generated scenarios are "generating" more power during 12am to 1am. This indicates the generated scenarios are representing the overall distribution of given training set.

Moreover, considering the simulated maximum solar power output value is 16MW, when there are 5% of wind samples existed in the training set, the midnight power generation is below 0.3MW. This indicates our model is robust to out-of-distribution bad data.

### D. Spatial Correlation

Instead of feeding a batch of sample vectors $x^{(i)}$ representing a single site's diurnal generation profile, here we feed GANs with a real data matrix $\{x^{(i)}\}$ of size $N \times T$, where $N$ denotes the total number of generation sites, while $T$ denotes the total number of timesteps for each scenario. Here we choose $N = 24, T = 24$ with a resolution of 1 hour. A group of real scenarios $\{x^{(i)}\}$ and generated scenarios $\{G(z^{(i)})\}$ for the 24 wind farms are ploted in Fig. 8. By visual inspection we find that the generated scenarios retain both the spatial and temporal correlations in the historical data (again, not seen in the training stage).

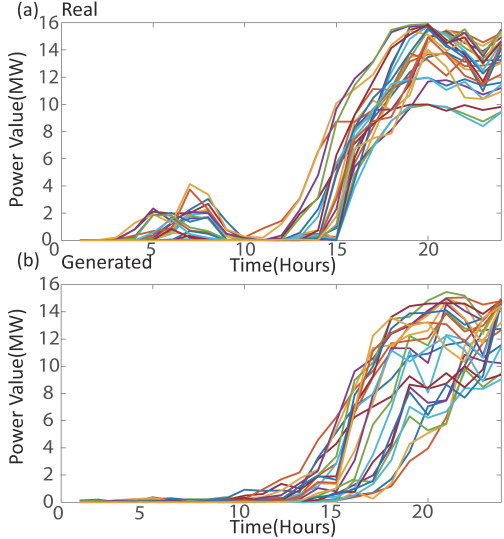

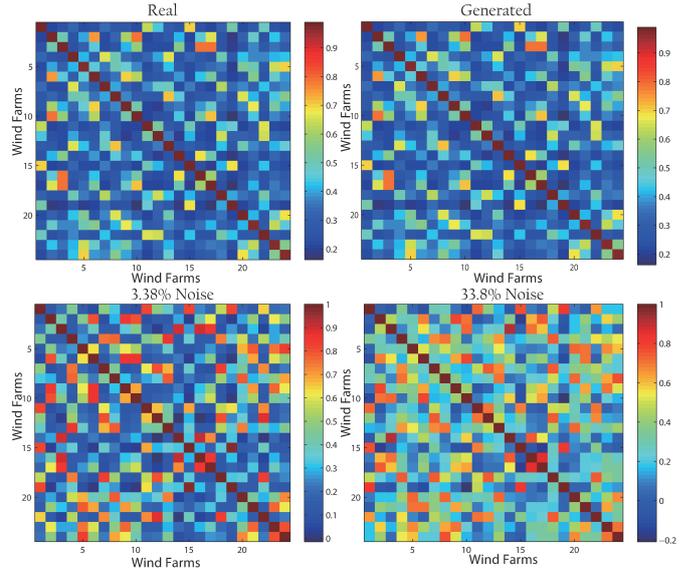

Fig. 9. The spatial correlation coefficients matrix colormap for a group of 24 wind farm one-day training scenarios on clean training data (Fig. 9a), scenarios generated using clean training data, scenarios generated using training data with 1% noise, and scenarios generated using training data with 10% noise.

Fig. 8. A group of one-day training (top) and generated (bottom) wind farm power output. The latter behaves similarly to the former both spatially and temporally.

To further validate the quality of generated scenarios, we add Gaussian white noise with standard deviation of 0.01 and 0.1 into the training data. The corresponding noise to signal ratio is of approximately 3.38% and 33.8%, respectively.[3] We compute the spatial correlation coefficients and visualize them in Fig. 9. Results show that generated scenarios' spatial correlation agrees with training sets, even under complex spatial correlation patterns. Thus GANs is able to capture both the spatial and temporal correlations at the same time. Moreover, with moderate amount of noise, our proposed method still finds the spatial correlation of power generation samples, while very low signal to noise ratio will lead to poor qualities.

---

[3]Here we define the noise to signal ratio as the inverse of the signal to noise ratio (SNR). The signal to noise ratio is the power of signal over the power of the noise, which is commonly used as a metric to quantify the level of noise.

### E. Conditional Scenario Generation

In this setting we are adding class information for the GANs' training procedure, and the generated samples could be "conditioned" to certain context or statistics informed by such labels. Here we present four representative applications using class information. For all these practical conditional scenario generation applications, we observe the generated samples are realistic. And we illustrate in Fig. 10 the learned overall marginal distribution compared to validation set.

*1) Wind Power Mean Values:* For a wind data set, we first calculate the mean value of power generation for each sample $x$, and classify these samples into 5 groups based on mean value $\mu(x)$ (MW): $\mu(x) < 0.5, \mu(x) < 1.5, \mu(x) < 3, \mu(x) < 6$ and $\mu(x) \geq 6$. Class information $y$ is encoded as an one-hot (indicator) vector. We then feed the new concatenated vector $(x, y)$ into GANs and train the model using Algorithm 1. We generate a group of 2,000 wind generation scenarios, with 400 samples in each month.

We evaluate these conditional generated samples by verifying the marginal distribution in the range of $[0MW, 16MW]$ for each class as shown in Fig. 10. The distribution value is divided into 10 equal bins. For each subclass of generated samples, it follows the same marginal distribution as the corresponding validation samples. For instance, within the class of $\mu(x) < 0.5$, generated samples' distribution informs us that it is unlikely to have large values of power generation. While in the class of $\mu(x) \geq 6$, over 30% of samples' generation value lies in the interval of $[14.4MW, 16MW]$, which can be used to simulate targeted high wind days. Yet for scenarios generated by Gaussian Copula method, though it can model the marginal distribution for wind power with smaller mean values (in class 1 and 2), it fails to represent the same marginal distribution for wind power with larger mean values. This is partially due to when sample mean generation value is larger, there are larger fluctuation and variability in realizations. Thus Gaussian Copula cannot represent the variability as well as our proposed method.

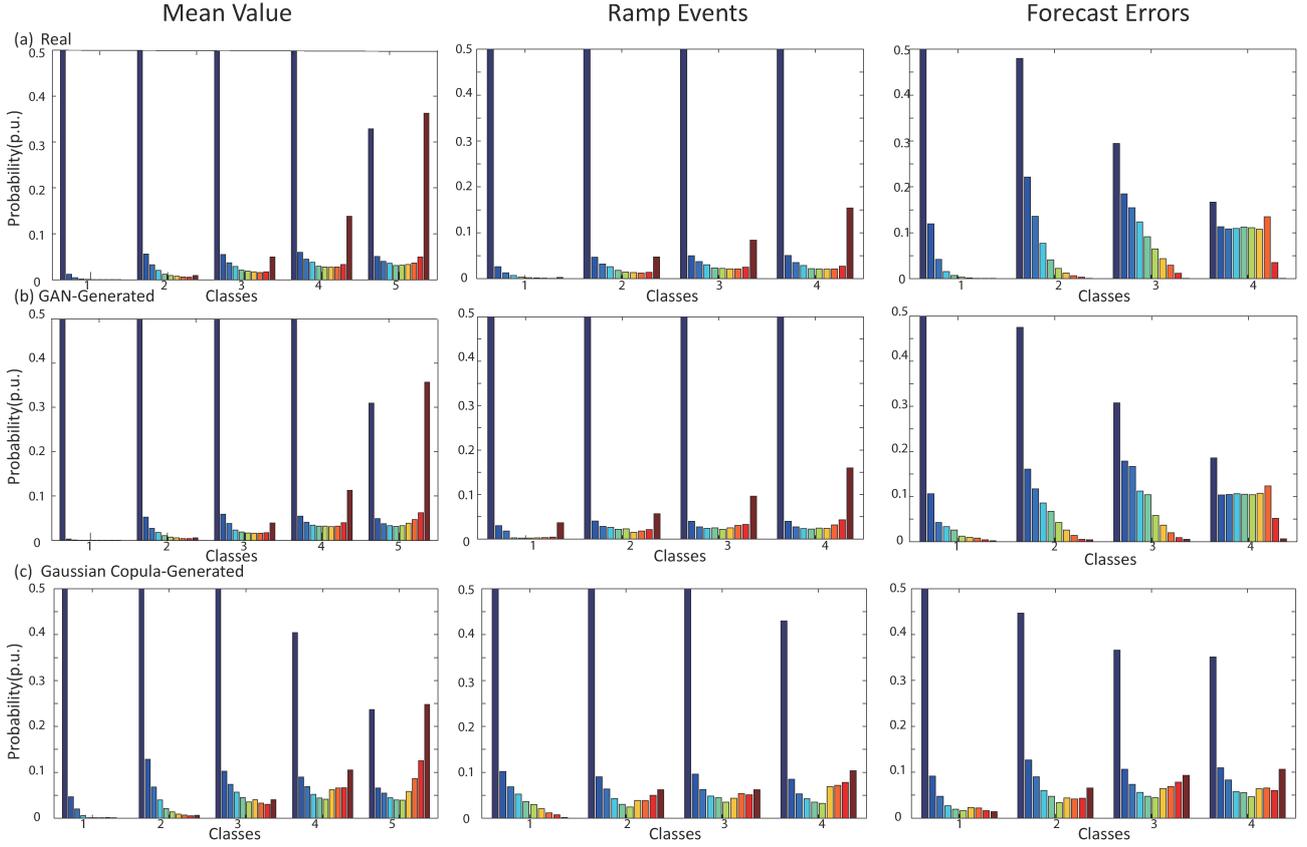

Fig. 10. Three group of conditional scenario generation for wind power comparing the marginal distribution of the real data, samples generated by GANs, and the samples generated by Gaussian copula. Three conditions, the wind power generations' mean value, 5-min ramp events, forecast errors level, are fed as conditional labels into our proposed model (Fig. 10b), and the generated scenarios' marginal distribution is compared with both the validation data (Fig. 10a) and scenarios generated by Gaussian copula method (Fig. 10c). In all cases, the marginal distribution of the samples generated by Gaussian copula tend to be more spread out than the actual marginal distributions and cannot accurately capture the extreme values. In contrast, the marginal distribution of the samples generated by GANs very closely resembles the real data distributions.

*2) Wind Power Ramp Events:* We further investigate GANs' model capabilities in generating wind scenarios conditioned on ramp events. Ramp events record the large fluctuations of renewables generation process, and generating scenarios for different ramp levels would help us characterize the fluctuating patterns and improve reliability of renewables generation. We examine if our proposed method could correctly capture the relationship between forecast values and the forecast errors. For each power generation sample $x$ of length $T$ with a resolution of 5 minutes, we define the generation ramp $\Delta(x)$ as the maximum absolute value of $30-$minute wind power changes:

$$\Delta(x) = \max(|x(t+30) - x(t)|), \quad t = 0, ..., T - 30 \quad (8)$$

We classify ramp events into 4 classes: 1. $\Delta(x) < 4.0MW$; 2. $4.0MW < \Delta(x) < 8.0MW$; 3. $8.0MW < \Delta(x) < 12.0MW$; 4. $12.0MW < \Delta(x) < 16.0MW$, and assign class labels to training scenarios. We feed the historical observations $x$ along with corresponding forecast error label into GANs, and show the simulated results of marginal distribution in the second column of Fig. 10. When there is an intense ramp existing (class 3 or 4), the wind samples have a smoother power generation distribution and generate more power than samples with small ramp levels (class 1 or 2). With same class of ramp events, generated scenarios follow the nearly same marginal distribution as the same class of validating forecast samples.

*3) Wind Power Forecast Errors:* In a similar conditional scenario generation setup for wind power mean value and ramp events, we examine if our proposed method could correctly capture the relationship between forecast values and the size of the forecast errors. Given a forecast, we break it into 4 classes depending on the size (total power) of the forecast. We feed the training forecast error vector along with forecast class label into GANs, and show the simulated results of marginal distribution in the last column of Fig. 10. Generated scenarios also follow the similar marginal distribution as the same class of validating historical samples. We can also observe that when forecast error is relatively large, both generated scenarios and original forecasts have larger power generation values. This indicates the difficulty in accurately forecasting wind power generation with larger output. As a comparison, forecast error scenarios generated by Gaussian copula method can not capture the overall distribution well.

In summary, since Gaussian copula method needs to explicitly estimate the training samples' copula, and uses a random number generator to realize scenarios, generated scenarios tend to fluctuate a lot in power values. Thus the scenarios' distribution are more scatted rather than concentrated in the same interval as the validating data in three groups of simulation.

In [33], the authors extend the generative model proposed in this paper to formulate an optimization problem to find a set of scenarios based on a given point forecast.

*4) Seasonal Solar Power Generation:* For the solar dataset, we add in labels based on month. So $y$ is a 12-dimension one-hot vector indicating which month the sample comes from. By adding this class information, we want to find out GANs is able to characterize the seasonal information reflected by $y$, e.g., a longer duration of power generation in the summer compared to the winter generation profile. Following the same training procedure as for conditional wind scenario generation, we generate a group of $2,400$ solar generation scenarios, with $200$ samples in each class.

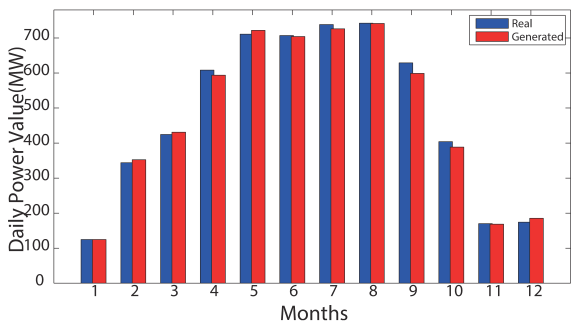

Fig. 11. Seasonal variation of daily power generation values for validating and generated solar power generation scenarios.

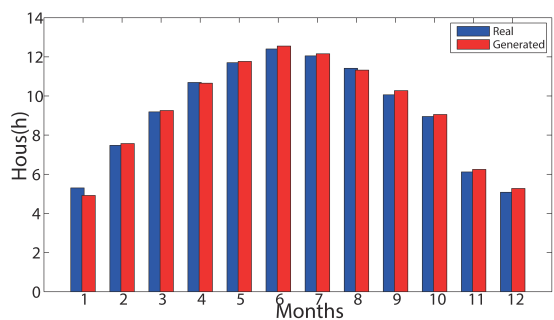

Fig. 12. Seasonal variation of daily power generation duration for validating and generated solar power generation scenarios.

To verify the generated samples indicate the seasonal patterns, we evaluate both the daily power generation sum values as well as daily power generation duration for each month's samples. The dry-summer Mediterranean climate existing in most parts of State of Washington is correctly identified by the generated samples, both based on power and duration. Fig. 11 shows the significant difference of daily power generation, while Fig. 12 also agrees with the seasonal variation of sunshine duration. With such scenarios generated based on months, power system operators are able to design seasonal-adaptive renewables dispatch strategies.

## V. Conclusion and Discussions

Scenario generation can help model the uncertainties and variations in renewables generation, and it is an essential tool for decision-making in power grids with high penetration of renewables. In this paper, a novel machine learning model, the *Generative Adversarial Networks (GANs)* is presented and proposed to be used for scenario generation of renewable resources. Our proposed method is data-driven and model-free. It leverages the power of deep neural networks and large sets of historical data to perform the task for directly generating scenarios conforming to the same distribution of historical data, without explicitly modeling of the distribution.

The case study using proposed model setup shows that GANs works well for scenario generation for both wind and solar. We also show in the simulation that by just retraining the model using historical data from multiple sites samples, GANs are able to generate scenarios for these sites with the corrected spatiotemporal correlations without any additional tuning. We also observe that by adding class information indicating scenario's properties, GANs is able to generate class-conditional samples conforming to the same sample properties. We validate the quality of generated samples by a series of statistical methods and compare results with Gaussican copula method for scenario generation.

Since our proposed methodology do not require any particular statistical assumptions, it can be applied to most stochastic processes of interest in power systems. In addition, as the method uses a feed forward neural network structure, it does not require sampling of potentially complex and high dimensional processes and can be scaled easily to systems with a large number of uncertainties. In future work, we propose to incorporate GANs into probabilistic forecasting problems. In addition, we also would like to utilize the scenarios and extend this work to the decision-making strategy design with high penetration of renewables generation.